\documentclass{article}

\usepackage{microtype}
\usepackage{bbding}
\usepackage{graphicx}
\usepackage{subfigure}
\usepackage{booktabs} % for professional tables
\usepackage{bm}
\usepackage{bbm}
\usepackage{pifont}  % 导入pifont宏包
\usepackage{hyperref}

\usepackage[accepted]{icml2024}

\usepackage{amsmath}
\usepackage{amssymb}
\usepackage{mathtools}
\usepackage{amsthm}

\usepackage[capitalize,noabbrev]{cleveref}

\theoremstyle{plain}
\newtheorem{theorem}{Theorem}[section]

\newtheorem{lemma}[theorem]{Lemma}

\theoremstyle{definition}
\newtheorem{definition}[theorem]{Definition}

\theoremstyle{remark}

\usepackage[textsize=tiny]{todonotes}

\icmltitlerunning{Submission and Formatting Instructions for ICML 2024}

\begin{document}

\twocolumn[
\icmltitle{Rethinking the Graph Polynomial Filter via \\Positive and Negative Coupling Analysis}




% \begin{icmlauthorlist}
% \icmlauthor{Bodong Du}{equal,yyy}
% \icmlauthor{Haodong Wen}{equal,yyy}
% \icmlauthor{Ruixun Liu}{yyy}
% \icmlauthor{Deyu Meng}{yyy}
% \icmlauthor{Xiangyong Cao\textsuperscript{$\dagger$}}{yyy}
% \end{icmlauthorlist}
\icmlsetsymbol{equal}{*}
\icmlsetsymbol{corresponding}{$\boldsymbol\dagger$}
\begin{icmlauthorlist}

\textbf{Bodong Du}\textsuperscript{$1*$}\quad
\textbf{Haodong Wen}\textsuperscript{$1*$}\quad
\textbf{Ruixun Liu}\textsuperscript{$1$}\quad
\textbf{Deyu Meng}\textsuperscript{$1$}\quad
\textbf{Xiangyong Cao}\textsuperscript{$1\dagger$}\\
\vspace{5pt}
\textsuperscript{$1$}Xianjiaotong University\quad
\end{icmlauthorlist}
% \icmlaffiliation{yyy}{Xianjiaotong University, Xian, Shanxi, China}
% \icmlcorrespondingauthor{Xiangyong Cao}{caoxiangyong@mail.xjtu.edu.cn}
\icmlkeywords{Machine Learning, Graph Neural Networks}

\vskip 0.3in
]
\let\thefootnote\relax\footnotetext{
\hspace*{-\footnotesep*3}
${}^*$Leading equal contribution.\\
${}^{\dagger}$Correspondence to: Xiangyong Cao (caoxiangyong@mail.xjtu.edu.cn).\\
\\preprint.
}



\begin{abstract}

Recently, the optimization of polynomial filters within Spectral Graph Neural Networks (GNNs) has emerged as a prominent research focus. Existing spectral GNNs mainly emphasize polynomial properties in filter design, introducing computational overhead and neglecting the integration of crucial graph structure information. We argue that incorporating graph information into basis construction can enhance understanding of polynomial basis, and further facilitate simplified polynomial filter design. Motivated by this, we first propose a Positive and Negative Coupling Analysis (PNCA) framework, where the concepts of positive and negative activation are defined and their respective and mixed effects are analysed. Then, we explore PNCA from the message propagation perspective, revealing the subtle information hidden in the activation process. Subsequently, PNCA is used to analyze the mainstream polynomial filters, and a novel simple basis that decouples the positive and negative activation and fully utilizes graph structure information is designed. Finally, a simple GNN (called GSCNet) is proposed based on the new basis. Experimental results on the benchmark datasets for node classification verify that our GSCNet obtains better or comparable results compared with existing state-of-the-art GNNs while demanding relatively less computational time. 

 %to effectively embed graph structure information into the basis construction process

\end{abstract}

\section{Introdution}
Graph Neural Networks (GNNs) are effective machine learning models for various graph learning problems~\cite{wu2022graph}, such as social network analysis~\cite{li2019encoding,qiu2018deepinf,tong2019leveraging}, drug
discovery~\cite{rathi2019practical,jiang2021could} and traffic forecasting~\cite{bogaerts2020graph,cui2019traffic,li2018diffusion}.Spectral Graph Neural Networks (GNNs) are a special kind of GNN that are designed from a signal-filtering perspective. Generally, the layer of spectral GNNs can be represented in a unified form~\cite{bridge}:
\begin{equation}\label{op1}
    \bm{H}^{(l+1)} = \sigma(\sum_s\bm{C}^{(s)}\bm{H}^{(l)}\bm{W}^{(l,s)}), 
\end{equation}
where $\bm{H}^{(l)}$ is the output of the $l^{th}$ layer, $ \sigma(\cdot)$ is the activation function, $\bm{W}^{(l,s)}$ is the weight to be learned, and $\bm{C}^{(s)}$ is the $s^{th}$ graph convolution support that defines how the node features are propagated to the neighboring nodes. Currently, most existing spectral GNNs differ from each other in selecting convolution supports $\bm{C}^{(s)}$. These GNNs were designed by analyzing the properties of the filter from a spectral perspective, i.e., the filter can be viewed as a signal filter in the graph Laplacian spectrum \cite{li2018deeper}.  %Theoretically, if we can get the eigenvalue decomposition of the graph Laplacian matrix, we can then obtain the optimal filter.\textcolor{red}{reference} 
We can understand $\bm{C}^{(s)}$ as a function of the eigenvalues of the Laplacian matrix~\cite{GCN}. However, the time cost of the eigenvalue decomposition for the Laplacian matrix is unaffordable, and thus researchers apply various polynomial functions of the Laplacian matrix to approximate the filter~\cite{hammond2011wavelets}.

\begin{figure*}[htp]
    \centering
    \includegraphics[width=13.5cm]{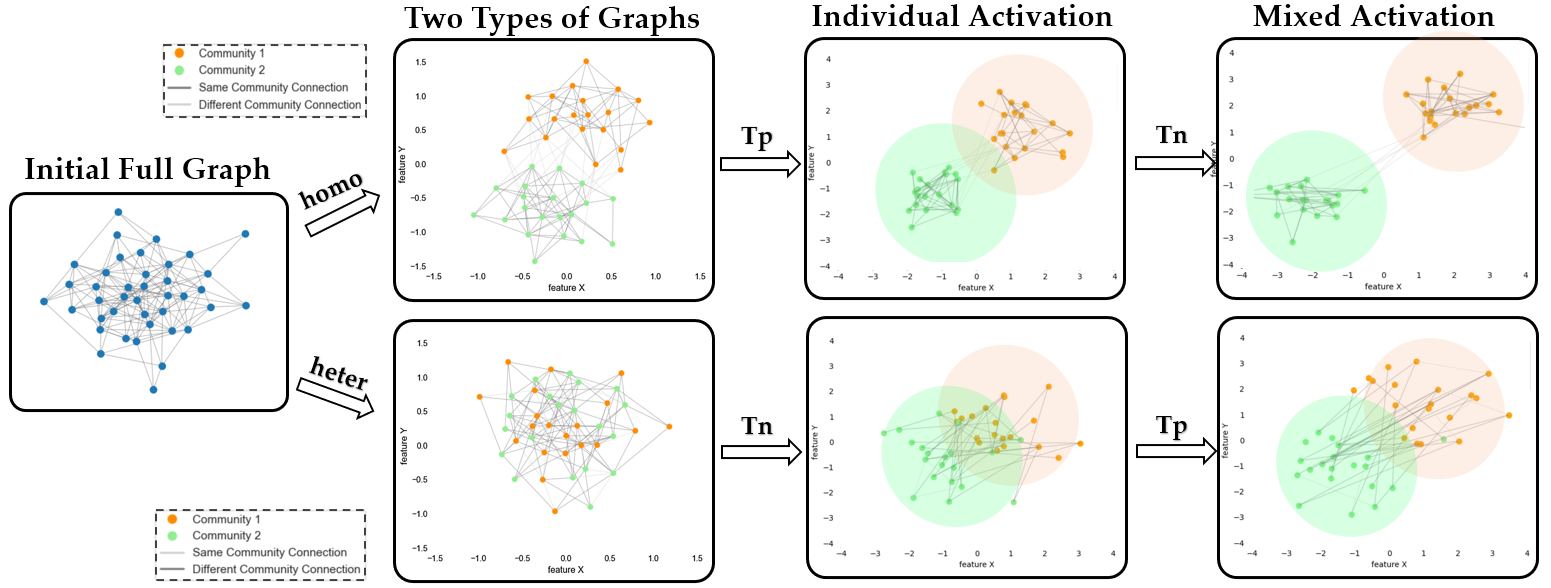}
    \caption{Effects of positive activation, negative activation and mixed activation on the feature distribution of the homophily and heterophily graphs. Under the setting of CSBM~\cite{deshpande2018contextual}, we generate an initial graph as shown on the left, then we further generate homophily and heterophily graphs with two different classes in the middle. For the homophily graph, we first tried positive activation, and representation distributions of different classes become apart. After using the negative activation, the two classes are more distinctly separated. The results are the same for the heterophily graph. } 
    \label{figure1}
\end{figure*}

Specifically, many researchers focus on analyzing the properties of polynomial filters and try to find the optimal basis from the existing polynomials to approximate the filter. For example, ChebNet~\cite{ChebNet} first adopts an orthogonal Chebyshev basis to approximate the filter. BernNet~\cite{BernNet} applies a non-negative Bernstein basis to improve the interpretation of GNN. JacobiConv~\cite{Jocobi} utilizes the orthogonal Jacobi basis due to its flexible approximation ability and faster convergence speed. OptBasisGNN~\cite{OptBasis} learns the optimal basis from orthogonal space. These spectral GNNs consider the properties of polynomials in an optimization manner and apply many complicated tricks to implement these bases, introducing additional computation costs. For example, BernNet~\cite{BernNet} has a quadratic time complexity related to the degree of the polynomial, JocobiConv~\cite{Jocobi} requires three iterative calculations to obtain the orthogonal basis, and OptBasisGNN~\cite{OptBasis} introduces an additional orthogonalization process to construct an orthogonal matrix for the advantage of optimization. In summary, current spectral GNNs mainly focus on utilizing the properties of polynomials to construct the basis, which not only introduces high computational costs but also neglects to fully exploit the graph information in the basis construction process. 

In this paper, we argue that utilizing the graph structure information in the basis construction process can help to understand the polynomial basis and simplify the design of polynomial filters, thus further inspiring us to construct graph polynomial filters with high accuracy and low computational cost. To achieve this goal, we propose a Positive and Negative Coupling Analysis (PNCA) framework to analyze and rethink polynomial filtering. Specifically, in this framework, we define the concepts of positive and negative activation on node and graph and illustrate the respective and mixed effects of positive and negative activation on the feature distribution for the homophily and heterophily graphs in Fig.~\ref{figure1}, which shows that individual positive activation and negative activation can separate different categories in the homophily graph and heterophily graph in the feature space, respectively. Besides, mixed activation can further strengthen the separation property of the graph. Moreover, we explore the PNCA from the message propagation perspective, revealing the subtle information hidden in the activation process. 
%The PNCA   intricately dissects key components of GNN, e.g. nodes, graph activation, and activation properties, providing a meticulous understanding of their roles within the broader context of graph optimization. 

Based on PNCA, we analyzed the main polynomial filters, e.g., GCN~\cite{GCN}, JKNet~\cite{xu2018representation}, and BernNet~\cite{BernNet}, considering their positive and negative graph activation properties, and found that GCN and JKNet perform poorly on heterophily graphs due to the lack of negative activation, and BernNet fuses positive and negative activation to achieve better performance, but fails to decouple positive and negative activation, leading to an increase in computational cost. In this paper, we thus attempt to design a new polynomial basis inspired by the PNCA to implement the decoupling of positive and negative activation. Additionally, the graph structure information is also embedded in the construction process of the new basis. Further, we build a simple GNN (i.e. GSCNet), which has the characteristics of positive and negative activations weights. Also, We analyze the expressive ability and computational cost of GSCNet from the perspectives of message propagation mechanism and graph optimization. To validate the efficacy of our proposed GSCNet, we conducted extensive experiments on benchmark node classification datasets. Experimental results verify that our GSCNet can achieve superior or comparable performance compared to other state-of-the-art GNNs while demanding relatively less computational time.

\section{Notations and Preliminaries}

\textbf{Graph Neural Networks.} We denote an undirected and unweighted graph $G$ with vertex set $V$ and edge set $E$ as $G = G(V,E)$, whose adjacency matrix is $\bm{A}$. And $\bm{\hat{A}}$ is the adjacency matrix of graph $G$ with self-loop. The symmetric normalized Laplacian matrix of $G$ is defined as $\bm{L}=\bm{I} - \bm{D}^{-1/2}\bm{AD}^{-1/2}$, where $\bm{D}$ is the diagonal degree matrix of $\bm{A}$. Given a graph signal $\bm{x}\in R^{n}$, where $n=|V|$ is the node number of the graph, the graph filtering operation is defined as $\sum_{k=1}^{K}w_{k}\bm{L}^{k}\bm{x}$, where $w_k$ is the weight. We denote $\bm{L = U\Lambda U^\top}$ as the eigen-decomposition of $\bm{L}$, where $\bm{U}$ is the matrix of eigenvectors and $\bm{\Lambda}$ =$diag[\lambda_1, ..., \lambda_n]$ is the diagonal matrix of eigenvalues. 

\textbf{Polynomial filters.} The GNN that designs graph signal filters in the Laplacian graph spectrum is called a spectral GNN. To avoid the eigenvalue decomposition of the graph Laplacian matrix, spectral GNNs apply various polynomial bases to approximate the Laplacian filter on the frequency domain of the graph, which is called the frequency polynomial filter. These frequency polynomial filters construct the $\bm{C^{(s)}}$ term in Eq. (\ref{op1}) in a uniform polynomial form as other GNNs do, which are uniformly called polynomial filters.

\section{Positive and Negative Coupling Analysis  }
\subsection{Basic concept}
Inspired by graph   
\cite{andrilli_hecker_2023,bronson_costa_saccoman_gross_2024} and message passing  ~\cite{Gilmer2017,spherical_message_passing}, we propose a Positive and Negative Coupling Analysis (PNCA) framework for Graph Information in this section. Generally, the main idea of our analysis is to view the message passing propagation as a coupling of two types of graph information, i.e. positive information and negative information. Before introducing our main theorem, we first introduce some necessary definitions and lemmas briefly. 

\begin{definition}
    (\textbf{Node activation}) \textit{For an undirected and unweighted graph, the ($K$-step) node activation $\bm{x}^*_t$ of a given node $t$ equals the linear combination of its feature vector $\bm{x}_t$ and all its one-step to K-step neighbors’ feature vectors. Mathematically, the ($K$-step) node activation $\bm{x}^*_t$ is defined as}
\begin{equation}
    {\bm{x}^*_t}=\alpha _{t} \bm{x}_t + \sum_{k=1}^K\sum_{s \in N_k{(t)}} \alpha_s \bm{x}_s,  \label{activ:equation} 
\end{equation} 
\textit{where $N_k(t)$ is the K-order neighbor of target node $t$, $\alpha_{s}$ and $\alpha_{t}$ ($\alpha_{t} \geq 0$) represents combination coefficients.}
\end{definition}

\begin{theorem}
    The node activation is permutation invariant, i.e., given any permutation matrix $\bm{P}\in\{0,1\}^{n\times n}$ and node $t$ with its K-order neighbors $\textbf{x}_{t}=(\bm{x}_t,\bm{x}_{s}), s\in N_k{(t)}$,
    \begin{eqnarray}
        \bm{P}f(\textbf{x}_{t})=f(\bm{P}\textbf{x}_{t}),
    \end{eqnarray}
     where $f(\textbf{x}_t)=\alpha _{t} \bm{x}_t + \sum_{k=1}^K\sum_{s \in N_k{(t)}} \alpha_s \bm{x}_s.$  
\end{theorem}
This theorem ensures that the node activation has the permutation invariant property and thus constitutes a rational form of node embedding representation\cite{fan2023generalizing}. The proof of this theorem can be found in Appendix~\ref{A:Appendix}.

\begin{definition}
   \textbf{(Positive node activation)} \textit{The ($K$-step) activation of node $t$ is positive if and only if $\{\alpha_{s}\geq 0, s\in N_k{(t)}\}$, with at least one $\alpha_{s}>0$ and $\alpha_{t}>0$. Otherwise, the ($K$-step) node activation is negative.}
\end{definition} 

\begin{definition}
    \textbf{(Graph activation)} \textit{For a graph $G(V,E)$ with $|V|=n$ and feature matrix $\bm{X}\in \mathbb{R}^{n\times d}$, where $d$ is the dimension of node feature, the activation of $G$ is $ \bm{X}^*=\bm{T}\bm{X}$, where $\bm{T}\in\mathbb{R}^{n\times n}$ is a transformation matrix and the $i_{th}$ row of $\bm{X}^*$ corresponds to the $i_{th}$ node in the graph.}
\end{definition}

\begin{definition}
    \textbf{(Positive graph activation)} \textit{$G(V,E)$ is a graph with self-loop, $|V|=n$, and $X$ is the feature matrix of $G$. Let $\bm{T}\in\mathbb{R}^{n \times n}$, where $\bm{T}_{ij}>0$ if $(v_i,v_j)\in E$ and $\bm{T}_{ij}=0$ otherwise. Then, $\bm{X}^* = \bm{T}^k\bm{X}$ is a positive ($K$-step) activation of graph $G$. Otherwise, the ($K$-step) graph activation is negative.}
\end{definition}
    
In this paper, we represent the positive graph activation and the negative graph activation as $\bm{T}_{p}$ and $\bm{T}_{n}$, respectively.

To find a specific form of positive and negative graph activation, we will refer to the following lemmas and theorems to answer this question.

\begin{lemma}
    \textit{The non-negative weighted sum of finite positive graph activation is still positive.}
    \label{lem3.2}
\end{lemma}
The proof of this lemma can be found in Appendix~\ref{A:Appendix}.

Based on Lemma \ref{lem3.2}, we can obtain the following theorem.
\begin{theorem}
Let $\bm{L}$ be the Laplacian matrix of graph $G$ and $\bm{I}$ be the identity matrix. Then, $\sum_{j=0}^k$ $\alpha_j$ $(2\bm{I}-\bm{L})^j X$,($\forall \alpha_j \geq 0$) is a positive graph activation and $(\sum_{j=0}^{K_2}\beta_j \bm{L}^j)\bm{X}$ is a negative graph activation. 
\end{theorem}
Theorem 3.7 provides a way to set the form of positive and negative graph activation, which will be used to construct our basis. The proof of this theorem can be found in Appendix~\ref{A:Appendix}.

\subsection{Effect of Positive and Negative Activation}
%The definition of positive activation is the starting point of PNCA. We find a simple but effective definition of positive activation, this has a strong relation with homophily graph and nonhomophily graph, which we will fully discuss this issue later in this section.

%This stability aligns with the goal of achieving consistent and reliable activations across the graph.
In this section, we mainly focus on discussing the relationship between positive (negative) activation and homophily (heterophily) graph by analyzing the effect of positive and negative activation.

First, we explain Definition 3.3. In Definition 3.3, $\alpha_s > 0$ represents the concept of positive activation among neighboring nodes. This activation mechanism implies an influence loop where node activations mutually reinforce each other~\cite{wu2021learning}. $\alpha_t > 0$ means that an activated vertex retains its own feature within the updated information. This characteristic resembles a residual connection~\cite{chen2020simple,chen2020revisiting} within graph activation, promoting stability and coherence in the activation process. 

Second, as defined in Eq. (\ref{activ:equation}), activation of a node can be seen as the linear reconstruction of its neighbors and itself, and activation of a whole graph can be seen as a special linear transformation of features~\cite{hosseini2017community}. Moreover, the combination can be divided into positive and negative types. In the rest of this section, we mainly focus on the polynomial forms of this transformation. The novelty is that this time, we are not trying to approximate some spectral filter in polynomials. On the contrary, we attempt to construct a polynomial filter from a PNCA perspective. 

Besides, the message passing propagations tend to push the nodes in the same concentrate around the class mean in the final hidden layer, the empirical property is called Neural Collapse (NC) in Graph Neural Networks~\cite{kothapalli2023neural,wu2022non}. Generally, the graph can be divided into two categories, i.e. homophily graph and heterophily graph. In a homophily graph, the nodes close to each other can be the same class with high probability. There, by combining the finding in the homophily graph and NC phenomenon, we guess that the optimal representation of a message passing propagation (or polynomial filter ) in a homophily graph is to learn a positive combination of its neighborhood, and the neighborhood thus tends to converge to a dense cluster, and the neighborhood can be accordingly in the same class with high probability. Here the word "positive" means that all nodes in the neighborhood should contribute a positive leading coefficient ahead of the node's feature vector. It's easy to see that the positive here perfectly matches the above definition of positive activation. In summary, for a homophily graph, the optimal polynomial filter tends to be positive. A similar conclusion can also be made in the heterophily graph that the optimal polynomial filter tends to be negative. 

%(Graph Structure Combined Net)

To further understand the properties of mixed activations, we introduce a remark as follows:

\textbf{\textit{Remark:}} The weighted sum of positive and negative graph activations is beneficial for the fusion of graph information, and thus may enhance the expressive capability of GNN and alleviate the over-smoothing tendencies.

To clarify this remake, we provide a rough analysis using the Unified Graph Optimization Framework~\cite{zhu2021interpreting}, which is denoted as 
\begin{equation}
   \min_{\bm{Z}} { \zeta \Vert \bm{F_1}\bm{Z}-\bm{F_2}\bm{H}\Vert_{\bm{F}}^2+\xi \text{tr}(\bm{Z}^\top \hat{\bm{L}}\bm{Z}) }. 
\end{equation}
Here, $\xi$ is a non-negative coefficient, $\zeta$ is usually chosen from $[0,1]$, and $\bm{H}$ is the transformation on original input feature matrix $\bm{X}$. $\bm{F_1}$ and $\bm{F_2}$ are defined as arbitrary graph convolutional kernels. $\bm{Z}$ is the propagated representation and corresponds to the final
propagation result when minimizing the objective $\bm{O}$.  Graph convolutional kernels $\bm{F_1}$ and $\bm{F_2}$ can be chosen
from the $\bm{I}$,$\bm{A}$, and $\bm{L}$, showing the all-pass, low-pass, and high-pass filtering capabilities, respectively. Designing polynomial filters is equivalent to designing $\bm{F_1}$ and $\bm{F_2}$. Additionally, we demonstrate the equivalence between concurrently designing $\bm{F_1}$ and $\bm{F_2}$ versus designing one over the other.

The weighted sum of positive and negative graph activations can be described by $\bm{F_1} = \bm{I}$ and $\bm{F_2} = \alpha \bm{T_p} +\beta \bm{T_n} $. Notably, $\bm{F_2}$ comprises a weighted sum of positive and negative activations. Next, we analyze $\bm{F_2}$ from two aspects. First, from the perspective of spectral domain expression, positive activation captures low-frequency information, while negative activation acquires high-frequency information. And their combination in $\bm{F_2}$ facilitates the integration of high and low-frequency information, thereby augmenting the model's expressive capacity across a broader spectrum~\cite{zhu2021interpreting}. Second, the flexibility of the weight coefficients (i.e. $\alpha$ and $\beta$ in $\bm{F_2}$) could enhance GNN's ability to mitigate over-smoothing compared to networks with fewer weighted degrees of freedom~\cite{chen2020simple}

\subsection{Explain PNCA via Message Propagation}

In the context of information aggregation, PNCA can be understood from two perspectives, i.e. information aggregation and information types.

\textbf{Information aggregation:} Common information aggregation strategies include concatenation, max-pooling, and LSTM-attention. etc~\cite{pmlr-v80-xu18c,jin2022raw,leng2021enhance}. In our case, we use a simple weight-sum approach for aggregation. 

\textbf{Information types:}  In the investigation of information propagation mechanisms, the exploration of inherent properties before aggregation has been limited~\cite{hou2022measuring}. PNCA introduces a more detailed form of information through node activation, which includes both positive and negative attributes. This enhanced granularity of information has the potential to advance the design of information aggregation mechanisms in graph neural networks (GNNs). To further understand the importance of this finer granularity, let us consider the formulation of node activation in PNCA. Denoting the activation of node $i$ at time step $t$ as $\bm{Z_i^t}$, PNCA computes this activation as a combination of positive and negative attributes, i.e.
 \begin{equation}
     \bm{Z_i^t}=\sum_{j\in N(i)} \underbrace{\alpha_{ij} \text{RELU}(\bm{T_p}\cdot \bm{X_j^t})}_{\text{Positive Message}} + \underbrace{\beta_{ij} \text{RELU}(\bm{T_n}\cdot \bm{X_j^t})}_{\text{Negative Message}}
     \label{meaasge}
 \end{equation}
Here, $N(i)$ denotes the neighborhood of node $i$. $\alpha_{ij}$ and $\beta_{ij}$ represent positive and negative attention weights. $\bm{X_j^t}$ is the feature representation of node $j$ at time step $t$. The first term in Eq.~(\ref{meaasge}) represents the Positive Message, while the second term represents the Negative Message. By incorporating both positive and negative attributes, PNCA introduces a nuanced perspective on information flow. This fine-grained approach enables PNCA to capture various aspects of information, leading to a more expressive representation of the node states. The positive attributes contribute positively to the activation, while the negative attributes help suppress irrelevant or contradictory information.

We use the label smoothness parameter $\lambda_l$ to measure the quality of surrounding information and smaller $\lambda_l$ means more positive information~\cite{hou2022measuring}. Then, we prove a theory to clarify the relationship between (positive and negative) activation and the label smoothness property.
\begin{theorem}
    Positive activation $\bm{T_p}$ in message propagation can decrease $\lambda_l$ and negative activation $\bm{T_n}$ can increase $\lambda_l$.
\end{theorem}
The proof of this theorem can be found in Appendix~\ref{A:Appendix}.

%\subsection{PNCA on popular bases}
%We present a comprehensive analysis of PNCA with a focus on popular bases to gain a thorough understanding of PNCA and its functionalities.

%Specifically, we delve into the analysis of Monomial $(1-\lambda)^ k$ and Bernstein $(2-\lambda)^{K-k}\lambda^ k$ bases. Our exploration, guided by PNCA  , reveals that the base $(\bm{2I}-\bm{L})^k$ exhibits  positive information activation but lacks a negative component, thus limiting its expressive power. Additionally, $(\bm{I}-\bm{L})^1$ lacks self-loops, rendering it negative.

%Moving forward, the base $(2\bm{I}-\bm{L} )^{K-k}\bm{L}^ k$ encompasses both positive and negative information activations. However, it fails to decouple the learning process of positivity and non-positivity. Consequently, achieving independent optimization of positive or negative activation within our network becomes unattainable. The learned leading coefficient in the term $(2\bm{I}-\bm{L} )^{K-k}\bm{L}^ k$ scales both positive and negative components simultaneously. This prompts the exploration of a simpler approach to decouple these components within our activation process.

\begin{figure}[htp]
    \centering
    \includegraphics[width=8.3cm]{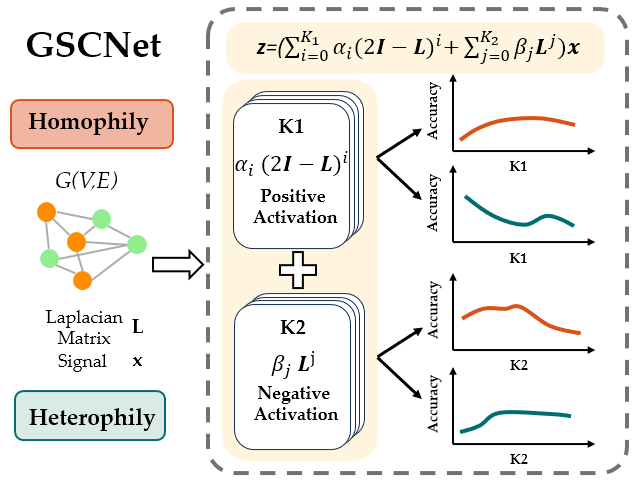}
    \caption{Effects of positive activation with degree $K_1$ and negative activation degree $K_2$ in our GSCNet. We evaluate each activation on both homophily graph (marked with orange) and heterophily graph (marked with green). The accuracy change with the degree parameters $K_1$ and $K_2$ are shown on the right.}
    \label{figure2}
\end{figure}

\section{Proposed Model}
In this section, we first analyze some basis in GNNs via PNCA. Then, we propose a new GNN based on a simple basis.

\subsection{Analysis on popular basis in GNNs via PNCA}
First, we conduct a comprehensive analysis of PNCA, focusing on popular bases to gain a thorough understanding of its functionalities. Specifically, we explore the Monomial $(1-\lambda)^ k$ and Bernstein $(2-\lambda)^{K-k}\lambda^ k$ bases. Our analysis, guided by PNCA, reveals that the base $(\bm{2I}-\bm{L})^k$ exhibits positive information activation but lacks a negative component, limiting its expressive power. Moreover, $(\bm{I}-\bm{L})^1$ lacks self-loops, resulting in negativity. Moving forward, the base $(2\bm{I}-\bm{L} )^{K-k}\bm{L}^ k$ incorporates both positive and negative information activations. However, it fails to decouple the learning process of positivity and non-positivity, preventing independent optimization. The learned leading coefficient in the term $(2\bm{I}-\bm{L} )^{K-k}\bm{L}^ k$ scales both positive and negative components simultaneously. This necessitates exploring a simpler approach to decouple these components within the activation process.

\begin{table*}
  \caption{Comparative Analysis between our GSCNet and Popular GNNs.}
  \label{tab:graph_matching}
  \centering
  \scalebox{0.9}{
  \begin{tabular}{@{}l|c|c c c}
    \toprule[1pt]
    Model    & Propagation Mechanism  & Positive activation & Negative activation & Decouple activation                           \\ \hline
GCN      & $\bm{Z} = \bm{A}^k\bm{X}\bm{W}^*$ \vspace{0.1em}
& \checkmark & \ding{55}   & \ding{55}    \\ \hline
JKNet    & $\bm{Z} = \sum_{k=1}^K \alpha_k \bm{A}^k\bm{X}\bm{W}^*$  \vspace{0.1em}      & \checkmark  &\ding{55}   &\ding{55}  \\ \hline
BernNet  & $\bm{Z} = \sum_{k=0}^K \alpha_k (2\bm{I}-\bm{L})^k\bm{L}^{K-k}\bm{X}\bm{W}^*$ \vspace{0.1em}     & \checkmark  &\checkmark   &\ding{55}  \\ \hline
GSCNet   & $\bm{Z} = (\sum_{i=0}^{K_1} \alpha_i{(2\bm{I}-\bm{L})^i}+\sum_{j=0}^{K_2}\beta_j \bm{L}^j)\bm{X}\bm{W}^*$ \vspace{0.1em} & \checkmark  &\checkmark   &\checkmark  \\ \hline
  \end{tabular}}
\end{table*}

\begin{table*}[h]
\centering
\caption{The statistics of our used real datasets for node classification.}
\tiny
\label{table3}
\resizebox{\textwidth}{!}{%
\begin{tabular}{l c c c c c c c c c c}
\hline
Datasets    & Cora & Computers & Photo  & Texas & Cornell & Citeseer & Actor & Chameleon & Pubmed & Penn94  \\ \hline
Nodes    & 2708 & 13752     & 7650   & 183   & 183     & 3327     & 7600  & 2277      & 19717  & 41554   \\
Edges & 5278 & 245861 & 119081 & 279   & 277     & 4552     & 26659 & 31371     & 44324  & 1362229 \\
Features & 1433 & 767       & 745    & 1703  & 1703    & 3703     & 932   & 2325      & 500    & 5       \\
Classes  & 7    & 10        & 8      & 5     & 5       & 6        & 5     & 5         & 5      & 2       \\ \hline
\end{tabular}%
}
\end{table*}

\begin{table*}[t]
\centering
\caption{Results on homophily graph datasets: mean accuracy (\%) ± 95\% confidence interval. The best results are in bold.}
\label{table_homo}
\resizebox{0.71\textwidth }{!}{%
\begin{tabular}{l c c c c c}
\hline
     & Computers  & Photo      & PubMed     & Citeseer   & Cora         \\ 
\hline
GCN         & 83.32±0.33 & 88.26±0.73 & 86.74±0.27 & 79.86±0.67 & 87.14±1.01   \\
ChebNet     & 87.54±0.43 & 93.77±0.32 & 87.95±0.28 & 79.11±0.75 & 86.67±0.82   \\
GPRGNN      & 86.85±0.25 & 93.85±0.28 & 88.46±0.33 & 80.12±0.83 & 88.57±0.69   \\
BernNet     & 87.64±0.44 & 93.63±0.35 & 88.48±0.41 & 80.09±0.79 & 88.52±0.95   \\
JocobiConvs & \underline{90.39±0.29} & \underline{95.43±0.23} & 89.62±0.41 & \underline{80.78±0.79}  & \underline{88.98±0.46}   \\
ChebNetII   & 89.37±0.38  &94.53±0.25  & 88.93±0.29 & 80.53±0.79 & 88.71±0.93 \\
OptBasis    & 89.65±0.25 & 93.12 ± 0.43  & \underline{90.30±0.19} & 80.58±0.82 & 85.02 ± 0.70     \\ 
\hline
GSCNet (ours)   & \textbf{90.85±0.30} & \textbf{95.87±0.31} & \textbf{91.16±0.34} & \textbf{80.92±0.56} & \textbf{89.26±0.38}   \\ \hline
\end{tabular}%
}
\end{table*}

\subsection{GSCNet}
 Based on the above analysis, we propose a new GNN, i.e. Graph Structure Combined Net (called GSCNet). To understand the GSCNet better, we first perceive it from the classic spectral-GNN perspective. Given an arbitrary filter function $h:[0, 2]\rightarrow[0, 1]$, the spectral filter on the graph signal $x$ is 
\begin{equation}
h(\bm{L})\bm{x}=\bm{U}h(\Lambda)\bm{U}^T\bm{x}=\bm{U}diag[h(\lambda_1), ..., h(\lambda _n)]\bm{U}^T\bm{x}.
\end{equation} 
In particular, given a graph signal $\bm{x}$, the convolution operator of our GSCNet in the Laplacian spectrum is 
\begin{equation}
    \bm{z}=(\sum_{i=0}^{K_1} \alpha_i{(2\bm{I}-\bm{L})^i}+\sum_{j=0}^{K_2}\beta_j \bm{L}^j)\bm{x}, \label{GSCNet}
\end{equation} 
where $K_1$ and $K_2$ are hyper-parameters, and $\alpha_i$ and $\beta_j$ denote the weight coefficients of the basis.

% \begin{table*}[!htbp]
% \centering
% \caption{\textbf{Experimental results.} of homophily datasets. Accuracies ± 95\% confidence intervals are displayed for each model on each dataset. The best performing two results are highlighted. The results of GCN, ChebNet, GPRGNN are taken from \cite{BernNet}. The results of BernNet, ChebNetII and
% JacobiConv and OptBasisGNN are taken from original papers. The results of GSCNet are the average of repeating experiments over 20 cross-validation splits.}
% \label{table4}
% \tiny
% \resizebox{0.8\textwidth    }{!}{%
% \begin{tabular}{l c c c c c}
% \hline
%      & Computers  & Photo      & Pubmed     & Citeseer   & Cora         \\ 
% \cmidrule(r){1-1}  \cmidrule(r){2-6}
% GCN         & 83.32±0.33 & 88.26±0.73 & 86.74±0.27 & 79.86±0.67 & 87.14±1.01   \\
% ChebNet     & 87.54±0.43 & 93.77±0.32 & 87.95±0.28 & 79.11±0.75 & 86.67±0.82   \\
% GPRGNN      & 86.85±0.25 & 93.85±0.28 & 88.46±0.33 & 80.12±0.83 & 88.57±0.69   \\
% BernNet     & 87.64±0.44 & 93.63±0.35 & 88.48±0.41 & 80.09±0.79 & 88.52±0.95   \\
% JocobiConv & \underline{90.39±0.29} & 95.43±0.23 & 89.62±0.41 & \underline{80.78±0.79}  & \underline{88.98±0.46}   \\
% ChebNetII   & 89.37±0.38  &94.53±0.25  & 88.93±0.29 & 80.53±0.79 & 88.71±0.93 \\
% OptBasis    & 89.65±0.25 & 93.12±0.43  & \underline{90.30±0.19} & 80.58±0.82 & 88.02±0.70     \\ 
% \cmidrule(r){1-1}  \cmidrule(r){2-6}
% GSCNet (ours)   & \textbf{90.55±0.30} & \textbf{95.57±0.31} & \textbf{91.16±0.34} & \textbf{80.92±0.56} & \textbf{89.26±0.38}   \\ \hline
% \end{tabular}%
% }
% \end{table*}
%%%%%%%%%%%%%%%%%%%%

\begin{table*}[t]
\centering
\caption{Results on heterophily graph datasets: mean accuracy (\%) ± 95\% confidence interval. The best results are in bold.}
\label{table_nonhomo}
\resizebox{0.71\textwidth}{!}{%
\centering
\begin{tabular}{l c c c c c}
\hline
     & Texas      & Cornell      & Actor        & Chameleon    & Penn94     \\ 
\hline
GCN         & 77.38±3.28 & 65.90±4.43   & 33.26±1.15 & 60.81±2.95   & 82.47±0.27 \\
ChebNet     & 86.22±2.45 & 83.93±2.13   & 37.42±0.58 & 59.51±1.25 & 82.59±0.31 \\
GPRGNN      & 92.95±1.31 & 91.37±1.81   & 39.91±0.62 & 67.49±1.37 & 83.54±0.32 \\
BernNet     & 93.12±0.65 & 92.13±1.64   & 41.71±1.12 & 68.53±1.68 & 83.26±0.29 \\
JocobiConvs & \underline{93.44±2.13} & \underline{92.95±2.46}   & 41.17±0.64  &  \underline{74.20±1.03}  & /       \\
ChebNetII   & 93.28±1.47  & 92.30±1.48  & 41.75±1.07 & 71.37±1.01 & \textbf{84.86±0.33} \\
OptBasis    & 88.62±3.44  & 83.77±4.75 & \underline{42.39±0.52} & \textbf{74.26±0.74}   & \underline{84.85±0.39} \\ 
\hline
GSCNet (ours)   & \textbf{96.22±0.98} & \textbf{94.59±1.31}   & \textbf{42.68±1.41}   & 72.09±1.64   & 82.64±0.43  \\ \hline
\end{tabular}%
}
\end{table*}

\textbf{\textit{Remark 1 (Analysis on Basis via PNCA):}} In Eq.~(\ref{GSCNet}), the proposed basis can be explained using our PNCA. $(2\bm{I}-\bm{L})^i$ represents positive activation of graph information. To adapt it to heterophily datasets, we introduce the $\bm{L}^j$ term as a negative activation. Essentially, GSCNet achieves information fusion by combining weighted mixtures of positive and negative activations. The weight parameters ($\alpha_i$ and $\beta_j$) are adaptable and learned by the model. Furthermore, the ratios of positive and negative activations can be adjusted by tuning $K_1$ and $K_2$. The activation illustrations within GSCNet are visually represented in Figure~\ref{figure2}. Further details on different degree parameters ($K_1$ and $K_2$) are discussed.  

\textbf{\textit{Remark 2 (Expressive Power Analysis on GSCNet):}} 
First, Table~\ref{tab:graph_matching} compares our GSCNet with popular GNNs, providing insight into their graph activation features and message propagation mechanisms. As can be seen, among these methods, only our GSCNet implements the positive and negative activation from the decoupling perspective. Additionally, GSCNet can enhance the expressiveness of GNNs by incorporating weighted sums of positive and negative graph information. Specifically, this approach assigns appropriate weights to positive and negative components, enriching the GNN representation~\cite{hou2022measuring}. The positive graph information activation $(2\bm{I}-\bm{L})^i$ positively contributes to the weighted sum, while the negative graph information activation $\bm{L}^j$ complements it. This combination captures a comprehensive understanding of the graph structure. By introducing weighted positive and negative graph information, GSCNet amplifies the GNN's expressive capacity, discerning both positive influences and critical negative signals. This innovative strategy fosters a context-aware representation, which is beneficial for GNNs.

From the unified graph optimization perspective, GSCNet is mathematically formulated as $\bm{F_1} = \bm{I}$ and $\bm{F_2} = \alpha(\bm{I}+\bm{\hat{A}})+\beta(\bm{I}-\bm{\hat{A}})$. The foundational $\bm{F_1} = \bm{I}$ preserves the initial graph representation, while $\bm{F_2}$ introduces a dynamic mechanism with weighted combinations of the identity matrix and adjacency matrix. This balance of positive and negative influences may mitigate the over-smoothing issue. The integration of $\bm{F_1}$ and $\bm{F_2}$ in GSCNet represents a sophisticated strategy that preserves the graph structure while strategically incorporating positive and negative influences. GSCNet thus may address the over-smoothing concerns within the unified graph optimization paradigm to some extent~\cite{zhu2021interpreting}.

\textbf{\textit{Remark 3 (Computational Complexity of GSCNet):}} 
GSCNet ensures computational efficiency through three key aspects: the closed-form multiplication property of matrix exponentiation, simplified weight aggregation, and the adoption of a graph convolutional kernel with O($dmf$) complexity~\cite{Jocobi}. These design choices collectively reduce complexity, making GSCNet well-suited for large-scale graph processing tasks.

\begin{table*}[!htbp]
\centering
\caption{Average running time per epoch (ms)/average total running time (s).}
\tiny
\label{table6}
\resizebox{1\textwidth}{!}{%
\centering
\begin{tabular}{l c c c c c c c}
\hline
            & Cora       & Citeseer  & Pubmed     & Computer   & Photo      & Texas    & Cornell   \\ 
\hline
GCN            & \underline{5.59/1.62}  & \underline{4.63/1.95}  & \underline{5.12/1.87} &  \underline{5.72/2.52} & \underline{5.08/2.63} & \underline{4.58/0.92}  & \underline{4.83/0.97}   \\
APPNP          & 7.16/2.32  &  7.79/2.77  &  8.21/2.63 & 9.19/3.48 & 8.69/4.18 & 7.83/1.63  &  8.23/1.68   \\
ChebNet        & 6.25/1.76  & 8.28/2.56   & 18.04/3.03 & 20.64/9.64 & 13.25/7.02 & 6.51/1.34  & 5.85/1.22   \\
GPRGNN         &  9.94/2.21  & 11.16/2.37  & 10.45/2.81 & 16.05/4.38 & 13.96/3.94 &  10.45/2.16  &  9.86/2.05   \\
BernNet        & 19.71/5.47  &  22.36/6.32  &  22.02/8.19 & 28.83/8.69 & 24.69/7.37 &  23.35/4.81  &  22.23/5.26   \\
JocobiConv &6.40/3.10  &6.30/3.00 &6.60/4.90 &7.30/4.80  &6.40/4.80  &6.60/3.40 &6.50/3.40 \\
OptBasis       & 19.38/7.57 & 13.88/4.53 & 39.60/15.79 & 16.16/13.49 & 13.25/10.59 & 14.07/8.61 & 13.03/8.16 \\  
\hline
GSCNet (ours)   & \textbf{9.98/1.03}    & \textbf{7.42/0.76}   &\textbf{5.14/1.82}    & \textbf{5.87/2.27}   &\textbf{5.30/2.31}    & \textbf{5.11/0.85}  & \textbf{3.46/0.89}   \\  \hline
\end{tabular}%
}
\end{table*}

\section{Experiment}
%\label{others}
In this section, we conduct experiments on real-world datasets to evaluate the performance of our proposed GSCNet. All experiments are carried out on a machine with an RTX 3090 GPU (24GB memory), Intel Xeon CPU (2.20 GHz), and 256GB of RAM. 

\subsection{Node classification on real-world datasets}

\textbf{Experimental Setup.} We evaluate the performance of our GSCNet on real-world datasets. Following ~\citep{BernNet} and ~\citep{OptBasis}, we adopt 5 homophily graphs, i.e., Cora, Citeseer, Pubmed, Computers and Photo, and 5 heterophily graphs, i.e., Chameleon, Actor, Texas, Cornell and Penn94 in our experiments. The statistics of these datasets are summarized in Table~\ref{table3}. We perform a full-supervised node classification task, where we randomly split each data set (except Penn94) into a train/validation/test set with a ratio of 60\%/20\%/20\%. For Penn94, we use the partitioned data sets given in~\citep{OptBasis}.  The experiment setup is the same as~\citep{BernNet}. More detailed settings can be found in Appendix~\ref{B:Appendix}.

In all the experiments, our proposed method is GSCNet, which are defined as equation \ref{GSCNet}. Besides, the competing methods are GCN~\citep{GCN}, ChebNet~\citep{ChebNet}, GPR-GNN~\citep{GPRGNN}, BernNet~\citep{BernNet}, JacobiConvs~\citep{Jocobi}, ChebNetII~\citep{he2022chebnetii}, OptBasisGNN~\citep{OptBasis}. The micro-F1 score with a 95\% confidence interval is used as the evaluation metric. The polynomial degree $K$ in other competing methods is established following their original papers. For our GSCNet, the degree parameters (i.e., $K_1$ and $K_2$) are set between 0 and 6.

\begin{figure*}[t]
        \centering
        \includegraphics[width=12.5cm]{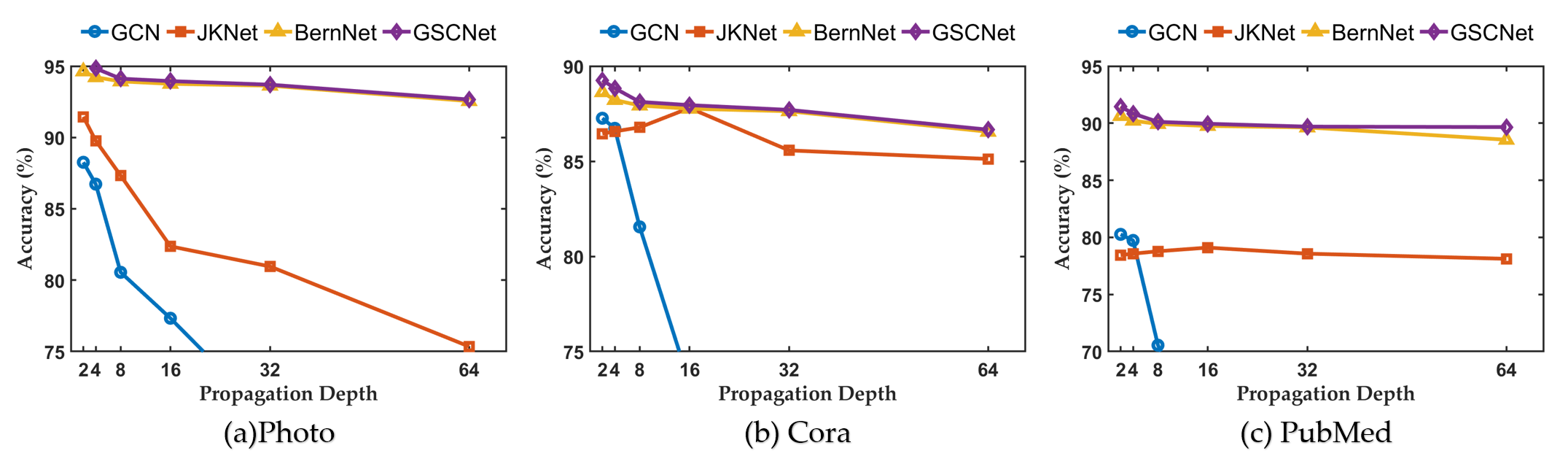}
        \caption{Comparison of different methods for alleviating the over-smoothing issue.}
        \label{Oversmooth}
\end{figure*}

\textbf{Results.} The experimental results on homophily graphs and heterophily graphs are summarized in Table \ref{table_homo} and Table \ref{table_nonhomo}. As can be seen, our GSCNet can consistently obtain the best performance on all five homophily datasets and three heterophily datasets (i.e., Texas, Cornell, Actor) and comparable results on Actor and Penn94 datasets compared with other SOTA methods.  

The training time for each method is shown in Table \ref{table6}. As can be seen, our proposed method has a lower average running time and a lower total running time than most other methods, and thus they should be more efficient and user-friendly in practice. This may be attributed to the following reasons. Firstly, since the multiplication operation is closed, i.e., $(2\bm{I}-\bm{L})^{k-1}*(2\bm{I}-\bm{L})=(2\bm{I}-\bm{L})^k$, the calculation of the monomial basis is thus simpler than the general polynomials. Specifically, the computational time complexity is $O(n)$ because of the cyclic computation of polynomials and reusability of $2\bm{I}-\bm{L}$ and $\bm{L}$. Secondly, since our basis does not require a big degree parameter, the parameter learning of our methods is relatively easier than other methods. More comparisons on running time can be found in Appendix~\ref{C:Appendix}.

\subsection{Node classification on Large-scale datasets}
\textbf{Experimental Setup.} We also conduct node classification experiments on two large heterophily graphs from the LINKX datasets~\cite{lim2021large}, i.e. Pokec and Wiki. For Pokec datasets, we use the five given splits to align with other reported experiment results. For Wiki dataset, since the splits are not provided, we use five random splits. For baselines, we choose MLP, GCN, GPRGNN, BernNet, ChebNetII, and OptBsisGNN. For more details, please refer to Appendix~\ref{C:Appendix}.

\begin{table}[h]
  \centering
  \caption{Experimental results on large-scale datasets: mean accuracy (\%) ± 95\% confidence interval/average running time (s).}
  \label{tab:result}
  \begin{tabular}{lccc}
    \toprule
    Dataset & Pokec &  Wiki \\
    $\Vert V\Vert$ & 1,632,803 &1,925,342\\
    $\Vert E\Vert$ & 30,622,564 &303,434,860\\
    \midrule
    MLP &  62.37 $\pm$ 0.02/24.6 &  37.38$\pm$ 0.21/31.6 \\
    GCN & / & / \\
    GPRGNN &  80.74 $\pm$ 0.22/35.7 &  58.73 $\pm$ 0.34/46.7 \\
    BernNet &  81.67 $\pm$ 0.17/483 &  59.02 $\pm$ 0.29/692\\
    ChebNetII &  82.33 $\pm$ 0.28/58.3 & 60.95 $\pm$ 0.39/70.4\\
    OptBasisGNN & 82.83 $\pm$ 0.04/---- & 61.85 $\pm$ 0.03/---- \\
    \midrule
    GSCNet &  81.97 $\pm$ 0.17/24.0 & 59.62 $\pm$ 0.29/31.2\\
    \bottomrule
  \end{tabular}
\end{table}

\begin{figure}[h]
\centering
\includegraphics[width=8cm]{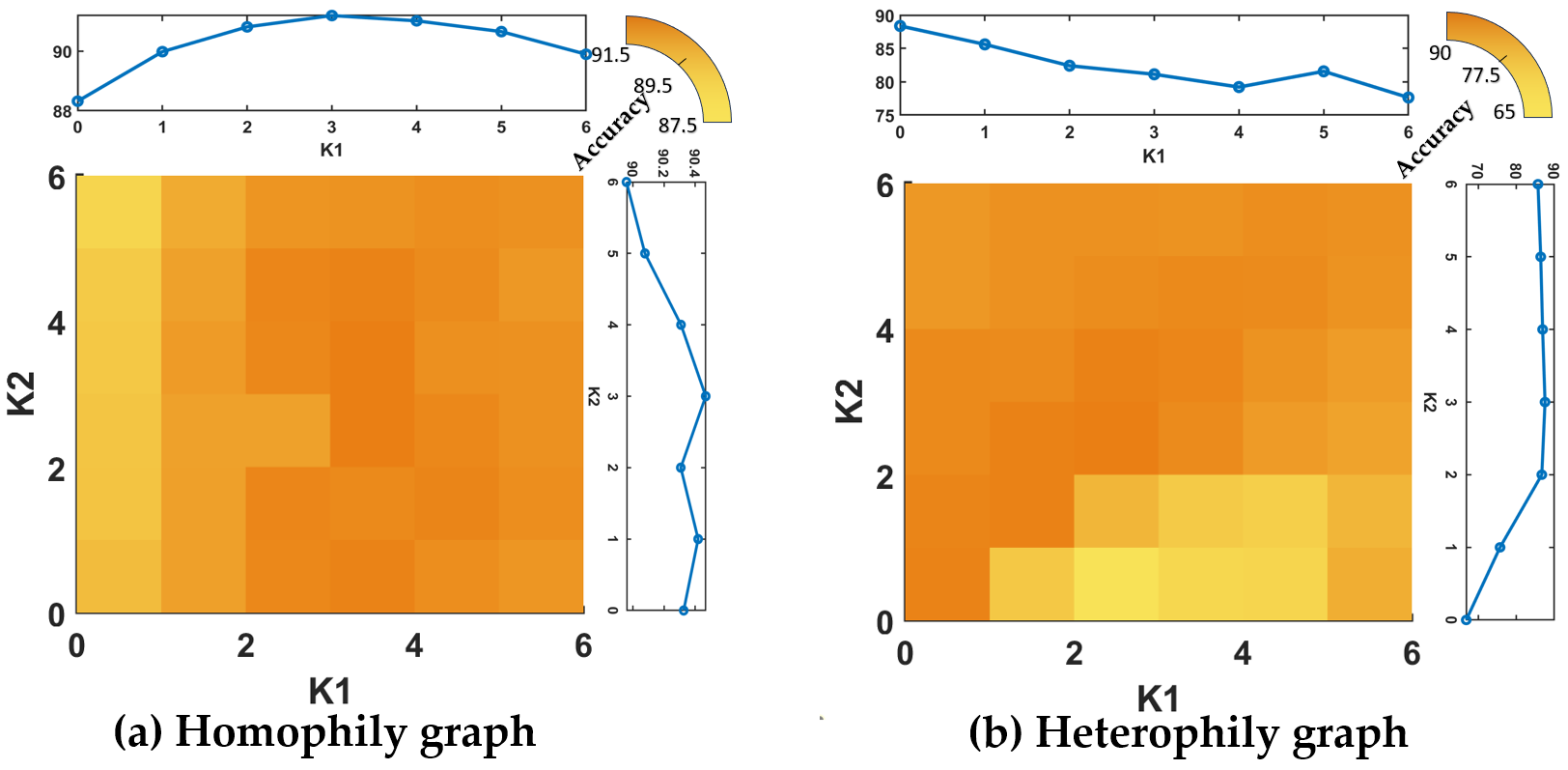}
\caption{Accuracy of different degree parameters ($K_1$ and $K_2$) for our GSCNet.}
\label{K1K2}
\end{figure}

\textbf{Results.} The experimental results are shown in Table~\ref{tab:result}. As can be seen, on Pokec and Wiki datasets, our proposed GSCNet can obtain comparable performance compared with the SOTA methods with less computational time. Therefore, our method may have a certain advantage over other models in terms of time on large-scale graph datasets.

\begin{figure}[h]
        \centering
        \includegraphics[width=6cm]{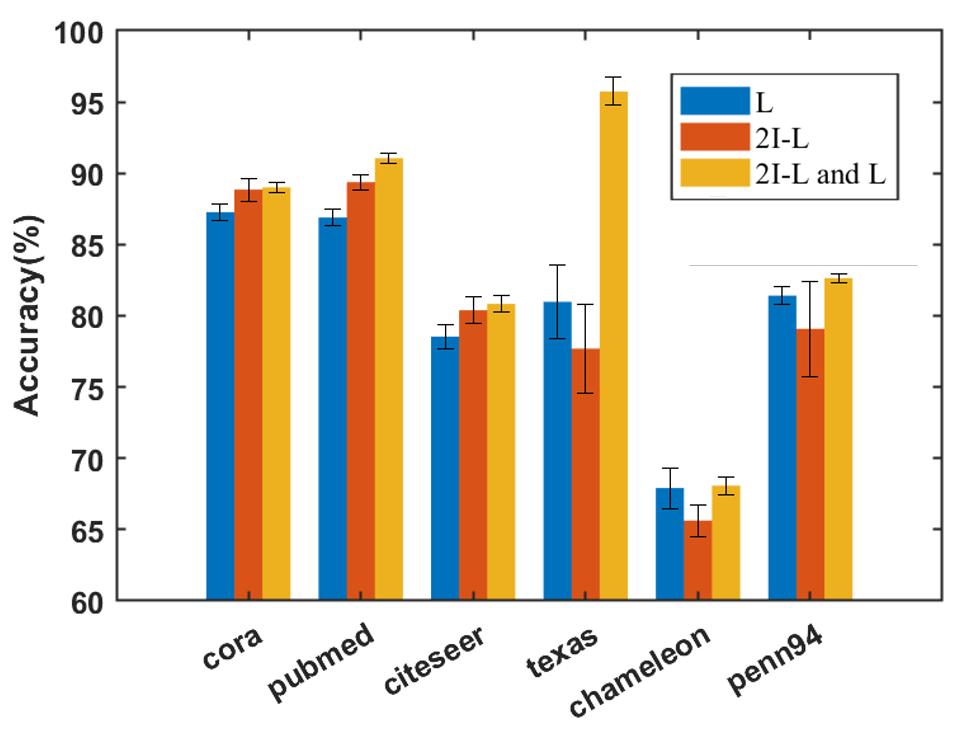}
        \caption{Performance comparison of different activations.}
        \label{figure3}
\end{figure}

\subsection{Discussions}
\textbf{Effect of mitigate over-smoothing issue.} To verify the superiority of our GSCNet in alleviating the over-smoothing issue, we experiment to compare our method with other three popular GNNs, i.e., GCN, JKNet, and BernNet, with the propagate depth increasing. The experimental results are shown in Figure~\ref{Oversmooth}. As can be seen, our GSCNet consistently exhibits the least decline in accuracy as the depth number increases, showcasing its superior ability to mitigate the over-smoothing issue. This resilience becomes particularly evident when compared to GCN, JKNet, and BernNet, highlighting GSCNet's robustness and nuanced balance between model complexity and generalization across varying depths of information propagation. 

% Empirical results on a computer dataset visually support these observations, emphasizing GSCNet's effectiveness in addressing over-smoothing challenges and providing valuable insights for enhancing the resilience of graph neural networks.

\textbf{Effect of setting different degree parameters ($K_1$ and $K_2$).} The results are shown in Figure~\ref{K1K2}. As shown in Figure~\ref{K1K2}(a), when $K_1$ is fixed and $K_2$ varies, the accuracy does not change so rapidly for
the homophily graph. When $K_2$ is fixed and $K_1$ varies, the accuracy changes relatively more rapidly. Generally, our GSCNet is not sensitive to the degree parameters on the homophily graphs. For the heterophily graph as shown in Figure~\ref{K1K2}(b), we can observe that our GSCNet is more sensitive to the parameters. No matter whether $K_1$ or $K_2$ is fixed, the accuracy will be affected obviously. Higher accuracy can be obtained when $K_1$ and $K_2$ are close.

\textbf{Effect of $K$-step graph activation’s properties.} 
To verify the effect of K-step activation's property, e.g. positive property, we further conduct a comparative experiment on three bases, i.e., $\sum_{i=0}^{K}(2\bm{I}-\bm{L})^i$, $\sum_{i=0}^{K}\bm{L}^i$ and $\sum_{i=0}^{K_1}(2\bm{I}-\bm{L})^i +\sum_{j=0}^{K_2}\bm{L}^j$. The experimental results are shown in Figure \ref{figure3}. As can be seen, due to its positive activation property, the basis $2\bm{I}-\bm{L}$ is more powerful in tackling the homophily graph while less effective in dealing with the heterophily graph compared to the negative basis $\bm{L}$. Furthermore, from Figure \ref{figure3}, we can observe that our GSCNet that mixed bases outperforms the two pure activation bases.

\section{Conclusion}
Considering that existing spectral GNNs often overlook the incorporation of crucial graph structure information during basis construction and incur computational overhead, we propose a Positive and Negative Coupling Analysis (PNCA) framework to effectively embed graph structure information into the basis construction process. Within this framework, we introduce concepts of positive and negative activation, exploring their individual and combined effects. Guided by PNCA, our analysis sheds light on mainstream polynomial filters. Inspired by PNCA, we design a novel basis that incorporates graph structure information during construction. This innovative approach informs the development of GSCNet, a concise GNN that achieves a refined balance between efficient computation and high precision. Furthermore, our GSCNet effectively alleviates the over-smoothing issue compared to popular GNNs. Experimental validations on real datasets demonstrate that GSCNet achieves better or comparable performance to existing state-of-the-art GNNs while requiring relatively less computational time.

\newpage

\bibliography{example_paper}
\bibliographystyle{icml2024}

%%%%%%%%%%%%%%%%%%%%%%%%%%%%%%%%%%%%%%%%%%%%%%%%%%%%%%%%%%%%%%%%%%%%%%%%%%%%%%%
%%%%%%%%%%%%%%%%%%%%%%%%%%%%%%%%%%%%%%%%%%%%%%%%%%%%%%%%%%%%%%%%%%%%%%%%%%%%%%%
% APPENDIX
%%%%%%%%%%%%%%%%%%%%%%%%%%%%%%%%%%%%%%%%%%%%%%%%%%%%%%%%%%%%%%%%%%%%%%%%%%%%%%%
%%%%%%%%%%%%%%%%%%%%%%%%%%%%%%%%%%%%%%%%%%%%%%%%%%%%%%%%%%%%%%%%%%%%%%%%%%%%%%%
\newpage
\appendix
\onecolumn

\section{Proof}
\label{A:Appendix}
\subsection{Permutation invariant}
\textbf{Thereom 3.2.}  The node activation is permutation invariant, i.e., given any permutation matrix $\bm{P}\in\{0,1\}^{n\times n}$ and node $t$ with its K-order neighbors $\textbf{x}_{t}=(\bm{x}_t,\bm{x}_{s}), s\in N_k{(t)}$,
    \begin{eqnarray}
        \bm{P}f(\textbf{x}_{t})=f(\bm{P}\textbf{x}_{t}),
    \end{eqnarray}
     where $f(\textbf{x}_t)=\alpha _{t} \bm{x}_t + \sum_{k=1}^K\sum_{s \in N_k{(t)}} \alpha_s \bm{x}_s.$   

\textbf{\textit{Proof:}} 
\begin{equation}
    \bm{P}f(\bm{x}_t) = \bm{P}(\alpha _{t} \bm{x}_t + \sum_{k=1}^K\sum_{s \in N_k{(t)}} \alpha_s \bm{x}_s) = \alpha _{t} \bm{P}\bm{x}_t + \sum_{k=1}^K\sum_{s \in N_k{(t)}} \alpha_s \bm{P}\bm{x}_s = f(\bm{P}x_t)
\end{equation}
\subsection{Positive graph activation}
\textbf{Lemma 3.6.} \textit{ The non-negative weighted sum of finite positive graph activation is still positive.}

\textbf{\textit{Proof:}} It can be proved because the sum of finite non-negative real numbers is still non-negative. 
\begin{equation}
    T_{sum} = \sum_{i=0}^n \alpha_i T_i
\end{equation}
\begin{equation}
    T_{sum_{ij}} = \sum_{k=0}^n \alpha_k T_{k_{ij}} 
\end{equation}
$\forall \alpha_k \geq 0, T_{k_{ij}} \geq 0$. So $T_{sum_{ij}} \geq 0$, then $T_{sum}$ is the positive graph activation.
\subsection{Specific form of graph activation}

\textbf{Theorem 3.7.} Let $\bm{L}$ be the Laplacian matrix of graph $G$ and $\bm{I}$ be the Identity matrix, then $\sum_{j=0}^k$ $\alpha_j$ $(2\bm{I}-\bm{L})^j X$, $\forall \alpha_j \geq 0$, is a proper and positive activation of $G$. 

\textbf{\textit{Proof:}} It can be proved that $(2\bm{I}-\bm{L})^j X$ is a  positive activation because of $(2\bm{I}-\bm{L})$ have the same form with matrix $\bm{T}$ in \textbf{Definition 3.5.}. And $\sum_{j=0}^k$ $\alpha_j$ $(2\bm{I}-\bm{L})^j X$ meets \textbf{Lemma 3.6}. So $\sum_{j=0}^k\alpha_j$ $(2\bm{I}-\bm{L})^j X$ is a positive activation of G. 

\subsection{Information type}
\textbf{Theorem 3.8.} Positive activation $\bm{T_p}$ in message propagation can decrease $\lambda_l$ and negative activation $\bm{T_n}$ can increase $\lambda_l$.

\textbf{\textit{Proof:}} First we introduce some definitions for our proof.
\begin{definition}
    (Label Smoothness). To measure the quality of surrounding information, we define the label smoothness as 
    \begin{equation}
        \lambda_l = \sum_{e_{v_i,v_j\in E}} (1-\mathbb{I}(v_i \simeq v_j))/\Vert E\Vert
    \end{equation}
\end{definition}

\begin{definition}
    (Information type)
    Smaller $\lambda_l$ means more positive information and bigger $\lambda_l$ means more negative information.
\end{definition}

When $\bm{T_p}$ apply to $\bm{x}$, the hidden representations of neighbor nodes will converge, resulting in close labels. Then $\mathbb{I}(v_i \simeq v_j)$ increase so the $\lambda_l$ becomes smaller so we prove the $Tp$ in message propagation is the positive information. Vice versa, $Tn$ is the negative information.

Our proof ends.

\vspace{0.5em}

\section{Details of Experiments}
\label{B:Appendix}

% \subsectionAppendix{Details In Experiments}
% \label{B:Appendix}

\subsection{\textbf{Model Setup.}}
For all the datasets, we followed the \citep{BernNet} structure. Especially in the two large-scale graph datasets, to enhance efficiency and alleviate CUDA overhead, we precompute crucial matrices, including the Laplacian matrix and some related matrices. These pre-computed values are subsequently directly utilized during the training process. For GSCNet, we initialized the weight parameters to uniform values of 1. We utilized a 2-layer MLP with 64 hidden units and optimized the learning rates for both the linear and propagation layers. Dropout was applied in both the convolutional and linear layers. Finally, we fine-tuned all parameters using the Adam optimizer.

\subsection{\textbf{Hyperparameter tuning.} }
We choose hyperparameters on the validation sets. We select hyperparameters from the range below with a maximum of 100 complete trials. For Chameleon, Actor and Penn94,  
\begin{itemize}
    \item  Truncated Order polynomial series: $K,K1,K2 \in 
   \{ 0,1,2,3,4,5,6 \}$ ;
   \item Learning rates for linear layer: \{ 0.008, 0.01, 0.012, 0.014, 0.02, 0.03, 0.04\}  ;
   \item Learning rates for propagation layer: \{ 0.01, 0.012, 0.014, 0.016, 0.0175, 0.02, 0.03, 0.04, 0.05\}  ;
   \item Weight decays: \{1e-8, · · · , 1e-3\};
   \item Dropout rates: \{0., 0.1, · · · , 0.9\};
\end{itemize}

For Wiki and Pokec,
\begin{itemize}
    \item Truncated Order polynomial series: $K,K1,K2 \in 
   \{ 0, 1, 2, 3, 4, 5, 6 \}$ ;
   \item Learning rates for linear layer: \{ 0.01, 0.012, 0.014, 0.018, 0.020, 0.022, 0.024, 0.03, 0.04\}  ;
   \item Learning rates for propagation layer: \{0.01, 0.015, 0.02, 0.025, 0.03, 0.04, 0.05 \}  ;
   \item Weight decays: \{0, 0.005\};
   \item Dropout rate: \{0., 0.05,0.1,...,0.9\}
\end{itemize}

For other datasets,
\begin{itemize}
    \item Truncated Order polynomial series: $K,K1,K2 \in 
   \{ 0, 1, 2, 3, 4, 5, 6 \}$ ;
   \item Learning rates for linear layer: \{ 0.01, 0.012, 0.015, 0.02, 0.022, 0.025, 0.03, 0.04, 0.05\}  ;
   \item Learning rates for propagation layer: \{0.01, 0.02, 0.022, 0.024, 0.03, 0.04, 0.05 \}  ;
   \item Weight decays: \{0, 0.005\};
   \item Dropout rate: \{0., 0.1, 0.2, 0.3\}
\end{itemize}

\subsection{Description of the source of some of the results}
 The results of GCN, ChebNet, and GPRGNN are taken from \cite{BernNet}. The results of BernNet, ChebNetII, JacobiConv, and OptBasisGNN are taken from original papers. The results of GSCNet are the average of repeating experiments over 20 cross-validation splits.

In particular, for Pokec and Wiki, due to their large-scale graph datasets, 'OOM' (Out of Memory) issues were encountered. The authors of ~\citep{OptBasis} did not publicly release the corresponding handling code, and it was not available in the paper. Therefore, we marked these instances with '--' to signify the unavailability of the information.

\section{Comparison of running time}\label{C:Appendix}
The comparison of running time for all the methods is shown in Figure 4-7. As can be seen, the area under the curve is the total running time of each dataset. Obviously, GSCNet's area is the smallest among other models which means that our model has extremely low computational overhead. Although the number of GSCNet's total running epochs is a bit high, its running time per epoch is short. It verifies the convergence rate analysis of GNN since our model's polynomial may not have a faster convergence rate but the computational cost of the polynomial itself is so low that it can compensate for the lack of convergence speed.  

\begin{figure}[htbp]
\centering
\begin{minipage}[T]{0.48\textwidth}
\centering
\includegraphics[width=8cm]{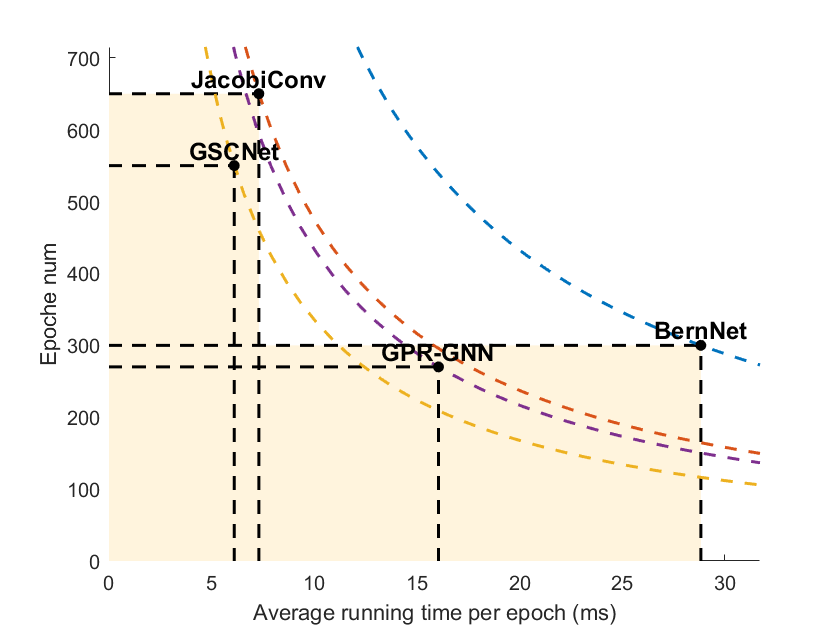}
   \caption{Running time on Computer dataset.}
\end{minipage}
\begin{minipage}[T]{0.48\textwidth}
\centering
\includegraphics[width=8cm]{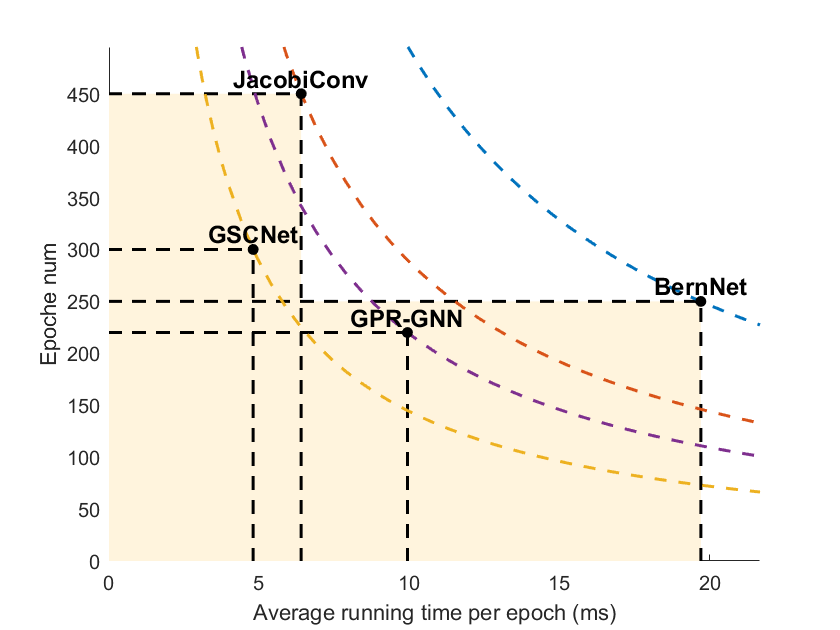}
   \caption{Running time on Cora dataset.}
\end{minipage}
\begin{minipage}[p]{0.48\textwidth}
\centering
\includegraphics[width=8cm]{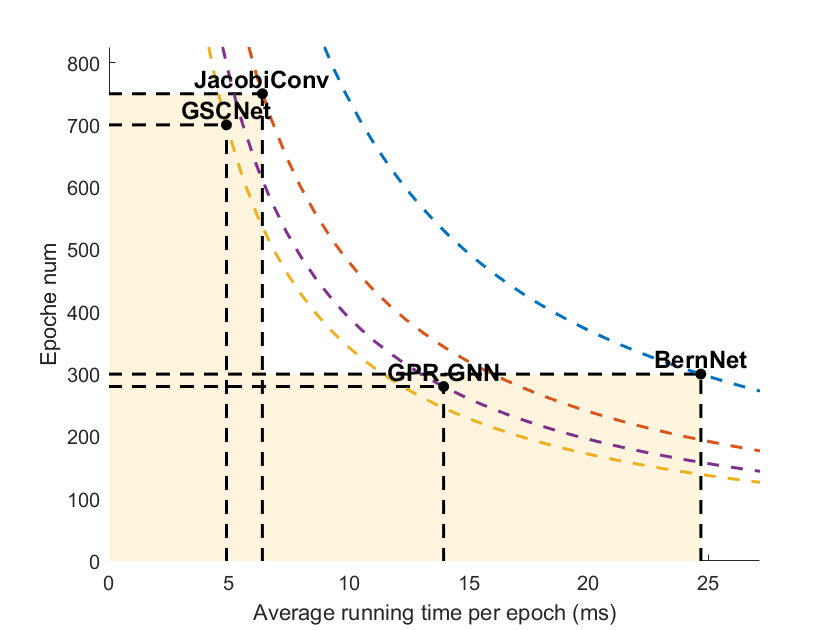}
   \caption{Running time on Photo dataset.}
\end{minipage}
\begin{minipage}[p]{0.48\textwidth}
\centering
\includegraphics[width=8cm]{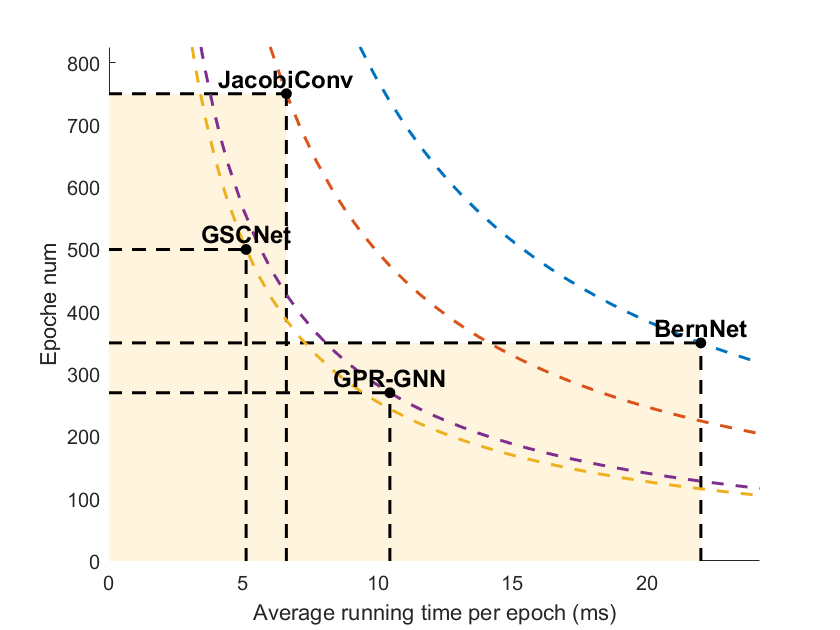}
   \caption{Running time on PubMed dataset.}
\end{minipage}
\end{figure}\label{f1}

\end{document}